\documentclass[]{article}
\usepackage[letterpaper]{geometry}
\usepackage{amta2020}
\usepackage{times}
\usepackage{url}
\usepackage{latexsym}
\usepackage{natbib}
\usepackage{layout}
\usepackage{booktabs}
\usepackage{todonotes}
\usepackage{xspace}
\usepackage[utf8]{inputenc}
\usepackage{footmisc}

\newcommand{\sockeye}{Sockeye\xspace}
\newcommand{\sockeyeTwo}{\sockeye~2\xspace}
\newcommand{\sockeyeURL}{\url{https://github.com/awslabs/sockeye}}
\newcommand{\mxnet}{MXNet\xspace}
\newcommand{\gluon}{Gluon\xspace}
\newcommand{\amt}{Amazon Translate}

\begin{document}

\title{\bf The Sockeye~2 Neural Machine Translation
\\
Toolkit at AMTA 2020}

\author{\name{\bf Tobias Domhan} \hfill \addr{domhant@amazon.com}\\
        \name{\bf Michael Denkowski} \hfill \addr{mdenkows@amazon.com}\\
        \name{\bf David Vilar} \hfill \addr{dvilar@amazon.com}\\
        \name{\bf Xing Niu} \hfill \addr{xingniu@amazon.com}\\
        \name{\bf Felix Hieber} \hfill  \addr{fhieber@amazon.com}\\
        \addr{Amazon}\\
        \name{\bf Kenneth Heafield}\footnotemark \hfill \addr{translate@kheafield.com}\\
        \addr{Efficient Translation Limited}
}

\maketitle
\pagestyle{empty}

\renewcommand{\thefootnote}{\fnsymbol{footnote}}
\footnotetext[1]{Work was done in an external advisory capacity.}
\renewcommand{\thefootnote}{\arabic{footnote}}

\begin{abstract}
  We present \sockeyeTwo, a modernized and streamlined version of the \sockeye neural machine translation (NMT) toolkit.
  New features include a simplified code base through the use of \mxnet's \gluon API, a focus on state of the art model architectures, distributed mixed precision training, and efficient CPU decoding with 8-bit quantization.
  These improvements result in faster training and inference, higher automatic metric scores, and a shorter path from research to production.
\end{abstract}

\section{Introduction}
\label{sec:introduction}

\sockeye \citep{hieber17:sockeye} is a versatile toolkit for research in the fast-moving field of NMT.
Since the initial release, it has been used in at least 25 scientific publications, including winning submissions to WMT
evaluations \citep{schamper18:WMT}.
\sockeye also powers \amt, showing industrial-strength performance in addition to the flexibility needed in academic environments.
Moreover, we are excited to see that hardware manufacturers are contributing to optimizing \mxnet \citep{chen2015mxnet} and \sockeye for speed.
Intel has demonstrated large performance gains for \sockeye inference on
Intel Skylake processors.\footnote{\url{https://www.intel.ai/amazing-inference-performance-with-intel-xeon-scalable-processors/#gs.wrgsji}}
NVIDIA is working on significant performance improvements for \sockeye's Transformer \citep{Vaswani:2017} implementation through fused operators and an optimized beam search.
This paper discusses \sockeyeTwo's streamlined \gluon implementation (\S\ref{sec:gluon}), support for state of the art architectures and efficient decoding (\S\ref{sec:modelling}), and improved model training (\S\ref{sec:training}).

\section{\gluon Implementation}
\label{sec:gluon}

\sockeyeTwo adopts \gluon, the latest and preferred API of \mxnet. \gluon simplifies the
code while improving overall performance.
Developers can define building blocks of neural network architectures as Python classes and seamlessly switch between eager execution for step-by-step debugging and cached computation graphs for maximum performance.
Migration to \gluon significantly simplifies training and inference code in \sockeyeTwo, reducing the overall number of lines of Python code by 25\%.
The hybridized \gluon transformer implementation in \sockeyeTwo improves training speed by 14\%, compared to \sockeye.

\section{Focus on State of the Art Models}
\label{sec:modelling}

\begin{table}
    \begin{center}
        \renewcommand\tabcolsep{4.5pt}
        \begin{tabular}{l|cc|c|cc|cc}
            & \multicolumn{2}{c|}{DE--EN} & EN--DE & \multicolumn{2}{c|}{FI--EN} & \multicolumn{2}{c}{EN--FI}  \\
            Layers & BLEU & Latency (ms) & BLEU & BLEU & Latency (ms) & BLEU & Latency (ms) \\
            \hline
            6:6  & 35.5 & 602 & 37.6 & 22.2 & 575 & 20.5 & 808 \\
            10:10 & 35.4 & 970 & --    & 22.3 & 863 & 20.8 & 1258 \\
            20:2  & 34.8 & 293 & 37.5 & 23.2 & 257 & 20.9 & 368 \\
        \end{tabular}
    \end{center}
    \caption{\label{tab:depth}SacreBLEU \citep{post:2018:WMT} scores and single-sentence latency on newstest2019 with varying numbers of encoder and decoder layers.  Latency values are the 90th percentile of translation time when translating each sentence individually (no batching).  We measure single sentence decoding latency on an EC2 c5.2xlarge instance with 4 CPU cores.  Except for EN-DE, we report the average over three independent training runs.}
\end{table}

Due to the success of self-attentional models, we concentrate development of \sockeyeTwo on the Transformer architecture \citep{Vaswani:2017}.
Our starting point is the ``base'' transformer with 6 encoder and decoder layers, model dimensionality of 512, and feed-forward layer size of 2048.
An exploration of different encoder and decoder depths shows that deep encoders with shallow decoders are competitive in BLEU and significantly faster for decoding.
Table~\ref{tab:depth} shows results with different numbers of encoder and decoder layers, denoted by $x$:$y$ where $x$ is the number of encoder layers and $y$ the number of decoder layers.
For FI-EN and EN-FI, the 20:2 model outperforms both the 6:6 model and the 10:10 model in terms of BLEU.
The 20:2 model also has roughly half the decoding latency of the 6:6 model and roughly one third the latency of the 10:10 model.
The relative efficiency of encoder versus decoder layers can be attributed to (1) the ability to parallelize across input tokens, (2) attention to only input tokens, and (3) not needing to run beam search on the source side.

\subsection{Source Factors}
\label{sec:factors}

\sockeye supports source factors in the spirit of \cite{sennrich-haddow-2016-linguistic}, additional representations that are combined with word embeddings prior to the first encoder layer.
In \sockeyeTwo, we improve source factor support by allowing different types of embedding combinations (concatenation, summation, or average), as well as weight sharing between source factor and word embeddings.

As an example application, we use source factors to represent input case.
Variations in case pose a challenge for machine translation systems as different orthographic variations are considered to be independent by the translation model (e.g., ``case'' is different from ``Case'' and both are different from ``CASE'').
We address these variations by lowercasing the input and encoding the original case information as a source factor (``lowercase'', ``capitalized'', ``all uppercase'' or ``mixed'').
We refer to this method as ``SF-case''.
An alternative is to lowercase and include the original cased word itself as a source factor, which we refer to as ``SF-word''.
As the original and lowercased versions of many words will be the same, it is useful to share the embeddings, a variant we refer to as ``SF-word-share''.

To evaluate the robustness of these strategies, we modify test sets by either entirely lowercasing, entirely uppercasing, or capitalizing the first character of each word.
We compare a baseline model that was trained on cased input (no source factors) against all ``SF-*'' methods.
The factored models also use BPE type factors as introduced by \citet{sennrich-haddow-2016-linguistic}.
Models use the 20:2 transformer architecture and training settings described in \S\ref{sec:modelling}.
Shown in Table~\ref{tab:case-robustness}, encoding case information with source factors is an effective way to improve robustness against case variation with the two versions of ``SF-case'' performing best.

\begin{table}
    \begin{center}
        \begin{tabular}{l|rrrr|rrrr}
            & \multicolumn{4}{c|}{DE -- EN} & \multicolumn{4}{c}{EN -- FI} \\
             & Ori & lower & Cap & UPP & Ori & lower & Cap & UPP \\
             \hline
            Baseline (cased)              & 36.7 & 33.0 & 22.9 & 9.1  & 20.7 & 16.4 & 3.4 & 1.1 \\
            SF-case (concat)     & 36.8 & 34.7 & 24.2 & 26.6 & 20.7 & 18.2 & 9.3 & 7.0 \\
            SF-case (sum)        & 36.8 & 34.8 & 23.0 & 28.4 & 21.4 & 18.3 & 7.7 & 7.2 \\
            SF-word              & 36.5 & 33.6 & 23.1 & 9.2 & 20.9 & 17.0 & 4.4 & 1.4 \\
            SF-word-share        & 36.8 & 33.8 & 21.9 & 9.3 & 21.4 & 17.2 & 4.0 & 1.3 \\
        \end{tabular}
    \end{center}
    \caption{Robustness results for several variants of representing case with source factors.  Models are evaluated on transformed versions of newstest2019: \textbf{Ori}ginal case, \textbf{lower}cased, \textbf{Cap}italization of the first character of each word, and \textbf{UPP}ERCASED. Scores are case-insensitive SacreBLEU \citep{post:2018:WMT}.}
        \label{tab:case-robustness}
\end{table}

\subsection{Quantization for Inference}
\label{sec:int8}

\sockeyeTwo now supports 8-bit quantized matrix multiplication \citep{quinn-ballesteros-2018-pieces} on CPUs based on the \texttt{intgemm} library.\footnote{\url{https://github.com/kpu/intgemm}}
By scaling values such that 127 corresponds to the maximum absolute value found in a tensor, matrix multiplication can be conducted with 8-bit integer representations in place of the default 32-bit floating-point representations without significant degradation of overall model accuracy.
Parameters can either be quantized offline and stored in a smaller model file or quantized on the fly at loading time.
Activations are quantized on the fly while other operators that consume far less runtime remain as 32-bit floats.

Latency-sensitive applications typically run with batch size 1 and small beam sizes, leaving little opportunity for batch parallelism.
Instead, matrix multiplication parallelizes over outputs of a layer.
To reduce latency, matrix multiplication and quantization are both parallelized with OpenMP.\footnote{\url{https://www.openmprtl.org}}
Layer outputs can be computed independently and the layer size is typically much larger than the batch size.
Parallelizing over layer inputs would require summing across threads.

Shown in Table~\ref{tab:quantization}, quantization significantly reduces non-batched decoding times with minimal effect on BLEU scores.
Improvement is most pronounced when running on a single CPU core while models using up to 4 cores still see a significant benefit.\footnote{For 1 and 2 cores, we set the number of OpenMP threads to 1 and 2 respectively.  For 4 cores, we set the number of OpenMP threads to 3 for best interaction with MXNet's own parallelization over operators.}

\begin{table}
    \begin{center}
        \begin{tabular}{lc|rrr|rrr}
            & & \multicolumn{3}{c|}{6:6 Layers} & \multicolumn{3}{c}{20:2 Layers} \\
            & CPUs & Time (s) & Tok/Sec & BLEU & Time (s) & Tok/Sec & BLEU \\
            \hline
            Baseline (fp32) & 1 & 1260.8 & 33.3 & 22.1 & 585.0 & 71.8 & 23.0 \\
            & 2 & 841.6 & 49.9 & 22.1 & 404.7 & 103.8 & 23.0 \\
            & 4 & 575.8 & 73.0 & 22.1 & 283.2 & 148.3 & 23.0 \\
            \hline
            Quantized (int8) & 1 & 511.6 & 82.1 & 22.0 & 285.7 & 147.0 & 22.8 \\
            & 2 & 435.9 & 96.4 & 22.0 & 242.3 & 173.4 & 22.8 \\
            & 4 & 334.3 & 125.7 & 22.0 & 173.0 & 242.9 & 22.8 \\
        \end{tabular}
    \end{center}
    \caption{\label{tab:quantization}CPU decoding times and SacreBLEU \citep{post:2018:WMT} scores for FI-EN newstest2019 with and without 8-bit quantization for both standard (6:6 layer) and deep encoder (20:2 layer) transformer models as described in \S\ref{sec:modelling}.  Models use a vocabulary selection shortlist of 200 items \citep{devlin17:intgemm} and translate one sentence at a time (batch size of 1).  Benchmarks are run on an EC2 c5.12xlarge instance (Cascade Lake processor) and limited to using 1, 2, or 4 CPU cores.}
\end{table}

\section{Training Improvements}
\label{sec:training}

\sockeyeTwo significantly accelerates training with Horovod\footnote{\url{https://github.com/horovod/horovod}} integration \citep{DBLP:journals/corr/abs-1802-05799} and \mxnet's automatic mixed precision (AMP).  Horovod extends synchronous training to any number of GPUs (including across nodes) while AMP automatically detects and converts parts of the model that can run in FP16 mode without loss of quality.
These methods also require additional computation per update (synchronizing data across distributed GPUs and checking reduced precision operations for overflow).
This overhead can be amortized by significantly increasing the effective batch size; gradients are aggregated per-GPU for several batches, then combined and checked for overflow for a single parameter update.
In practice, scaling the effective batch size by $N$, the learning rate by $\sqrt{N}$ \citep{krizhevsky2014one}, and leaving other hyper parameters unchanged works well for batches of up to 260K tokens.

\sockeye also provides a data-driven alternative to the popular ``inverse square root'' learning schedule used by \cite{Vaswani:2017} and \cite{ott-EtAl:2018:WMT}.
Termed ``plateau-reduce'', this scheduler keeps the same learning rate until validation perplexity does not increase for several checkpoints, at which time it reduces the learning rate and rewinds all model and optimizer parameters to the best previous point.
Training concludes when validation perplexity reaches an extended plateau.
In a WMT19 benchmark \citep{barrault-EtAl:2019:WMT}, plateau-reduce training produces stronger models in slightly less time than the setup described by \cite{ott-EtAl:2018:WMT}.
The results are presented in Table~\ref{tab:training} where all values are averages over 3 independent training runs with different random initializations and all models train until validation perplexity reaches a plateau.

The relevant hyper parameters for \sockeyeTwo's large batch training are an effective batch size of 262,144 tokens, a learning rate of 0.00113 with 2000 warmup steps and a reduce rate of 0.9, a checkpoint interval of 125 steps, and learning rate reduction after 8 checkpoints without improvement.
After an extended plateau of 60 checkpoints, the 8 checkpoints with the lowest validation perplexity are averaged to produce the final model parameters.
While Horovod enables scaling to any number of GPUs, we find that training on 8 GPUs on a single node still delivers the best value when considering both speed and cost.

\begin{table}
    \begin{center}
        \begin{tabular}{l|cc|cc}
            & \multicolumn{2}{c|}{DE--EN} & \multicolumn{2}{c}{EN--FI} \\
            & BLEU & Time & BLEU & Time \\
            \hline
            \cite{ott-EtAl:2018:WMT} & 34.7 & 30h & 20.1 & 14h \\
            Plateau-Reduce & \textbf{34.9} & \textbf{28h} & \textbf{20.7} & \textbf{12h} \\
        \end{tabular}
    \end{center}
    \caption{\label{tab:training}SacreBLEU \citep{post:2018:WMT} scores (newstest2019) and training times (8 NVIDIA V100 GPUs) for a 20 encoder 2 decoder layer transformer using the training setup described by \cite{ott-EtAl:2018:WMT} and plateau-reduce, both implemented in \sockeyeTwo.}
\end{table}

\section{Licensing and availability}
\label{sec:license}

\sockeyeTwo is available\footnote{\sockeyeURL} under the Apache 2.0 license.
It includes Docker builds to easily run training or inference with all of the latest features on any supported platform.

\section{Conclusion}
\label{sec:conclusions}

\sockeyeTwo provides out-of-the-box support for quickly training strong Transformer models for research or production.
Extensive configuration options and the simplified \gluon code base enable rapid development and experimentation.
As an open source project, we invite the community to contribute their ideas to \sockeyeTwo and hope that the new programming model and various performance improvements enable others to conduct effective and successful research.

\small

\bibliographystyle{apalike}
\bibliography{refs}

\end{document}